\documentclass[11pt,a4paper]{article}
\usepackage[hyperref]{naaclhlt2019}
\usepackage{times}
\usepackage{latexsym}
\usepackage{booktabs}
\usepackage{graphicx}
\usepackage{url}
\usepackage{array}

\usepackage{comment}
\usepackage{multirow}

\newcommand{\Secref}[1]{Section~\ref{#1}}
\newcommand{\secref}[1]{section~\ref{#1}}
\newcommand{\Tabref}[1]{Table~\ref{#1}}
\newcommand{\tabref}[1]{table~\ref{#1}}
\newcommand{\Figref}[1]{Figure~\ref{#1}}
\newcommand{\figref}[1]{figure~\ref{#1}}

\newcommand{\word}[1]{\emph{#1}}
\newcommand{\nlplabel}[1]{\textsc{#1}}

\definecolor{roamdarkblue}{HTML}{0499CC}
\definecolor{roamlightblue}{HTML}{03A9F4}
\definecolor{roamdarkgray}{HTML}{838A8A}
\definecolor{roamlightgray}{HTML}{B8B8B8}
\definecolor{roamgreen}{HTML}{4D8951}
\definecolor{roamblack}{HTML}{212121}
\definecolor{roamsteelblue}{HTML}{9BB8D7}
\definecolor{roamorange}{HTML}{FDBA58}
\definecolor{roamwhite}{HTML}{FAFAFA}
\definecolor{roampurple}{HTML}{876DB5}

\definecolor{superlightgray}{HTML}{DDDDDD}
\definecolor{superlightgreen}{HTML}{B4FFB4}
\definecolor{superlightorange}{HTML}{FFD090}

\newcommand{\roamdarkblue}[1]{{\color{roamdarkblue}#1}}

\newcolumntype{T}[1]{>{\raggedright\arraybackslash\hspace{0pt}}p{#1}}

\aclfinalcopy 

\title{Effective Feature Representation for Clinical Text Concept Extraction}

\author{
  Yifeng Tao \\
  Roam Analytics \\
  Carnegie Mellon University\\  
  \And
  Bruno Godefroy\\
  Roam Analytics\\
  \AND
  Guillaume Genthial \\
  Roam Analytics\\
  \And
  Christopher Potts\\  
  Roam Analytics\\
  Stanford University\\
}

\date{}

\begin{document}

\maketitle


\begin{abstract}
  Crucial information about the practice of healthcare is recorded
  only in free-form text, which creates an enormous opportunity for
  high-impact NLP. However, annotated healthcare datasets tend to be
  small and expensive to obtain, which raises the question of how to
  make maximally efficient uses of the available data. To this end, we
  develop an LSTM-CRF model for combining unsupervised word
  representations and hand-built feature representations derived from
  publicly available healthcare ontologies. We show that this combined
  model yields superior performance on five datasets of diverse kinds
  of healthcare text (clinical, social, scientific, commercial).  Each
  involves the labeling of complex, multi-word spans that pick out
  different healthcare concepts. We also introduce a new labeled
  dataset for identifying the treatment relations between drugs and
  diseases.
\end{abstract}


\section{Introduction}\label{sec:introduction}

The healthcare system generates enormous quantities of data, but its
tools for analytics and decision-making rely overwhelmingly on a
narrow subset of structured fields, especially billing codes for
procedures, diagnoses, and tests. The textual fields in medical
records are generally under-utilized or completely ignored. However,
these clinical texts are our only consistent source of information on
a wide variety of crucial factors -- hypotheses considered and
rejected, treatment rationales, obstacles to care, brand recognition,
descriptions of uncertainty, social and lifestyle factors, and so
forth. Such information is essential to gaining an accurate picture of
the healthcare system and the experiences of individual patients,
creating an enormous opportunity for high-impact NLP.

\begin{figure}[t]
  \centering
  \includegraphics[width=1.0\linewidth]{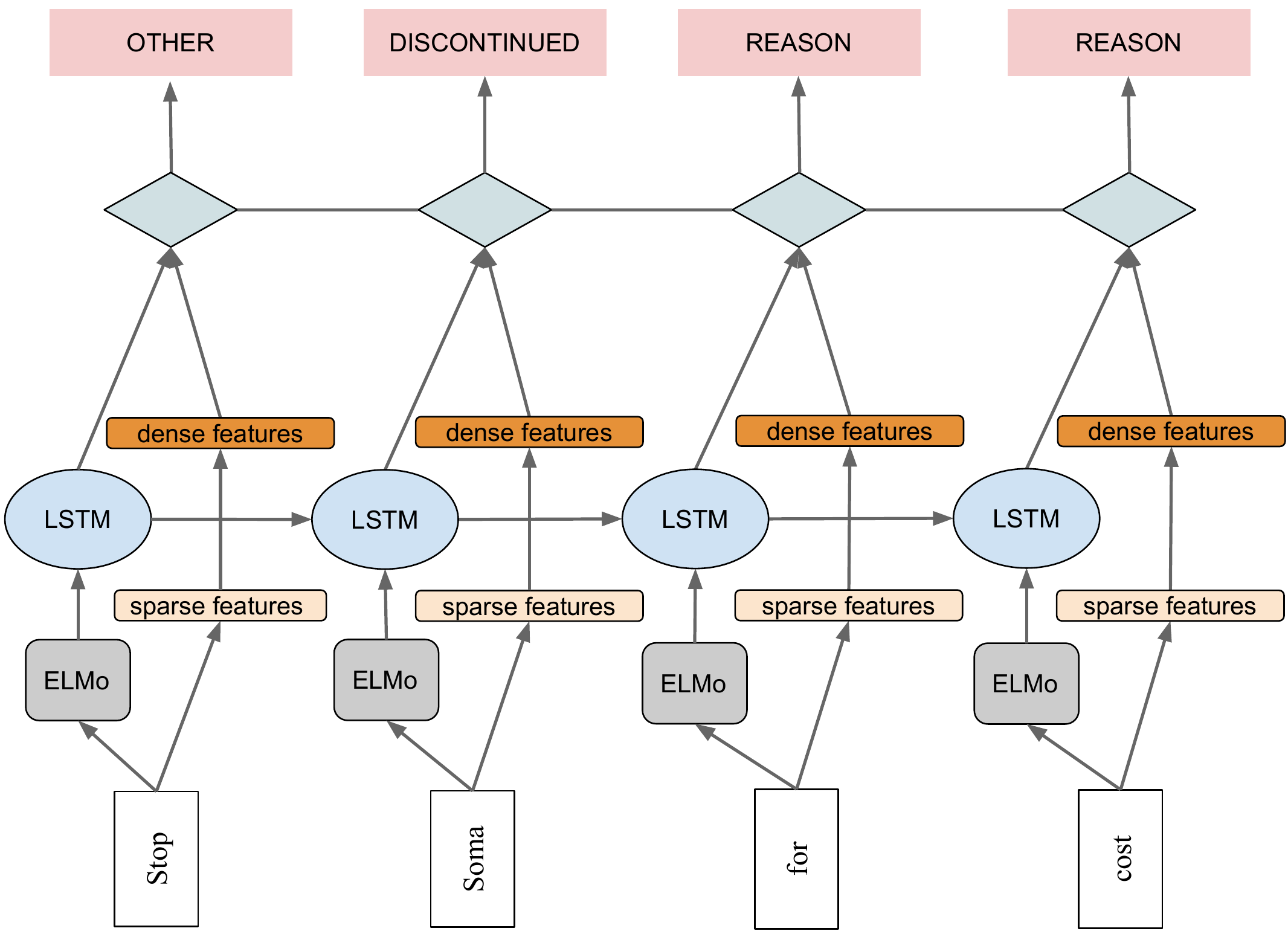}
  \caption{Model diagram. In our full model, words are represented by
    pretrained ELMo embeddings, which feed into LSTM cells,
    and by sparse ontology-derived feature representations, which are fed to a dense layer with dropout to
    produce a lower-dimensional representation that is
    concatenated with the hidden states of the LSTM. The resulting
    mixed feature representation is fed into a CRF layer
    that forms the basis for token-level label predictions. We assess this full model against variants without
    the LSTM or hand-built features to motivate the full version.}
  \label{fig:model}
\end{figure}

However, annotated clinical text datasets are scarce and tend to be
small, for two reasons. First, data access is usually highly limited
because of privacy considerations; the inherent richness of language
data means that de-identification is hard or impossible \cite{uzuner2007deidentification}. Second,
because healthcare concepts are complex, the needed annotations
generally must be provided by domain specialists who are trained both
in the practice of healthcare and in the interpretation of healthcare
records. Such experts are in high demand, and the annotation work they
do is intellectually challenging, so the annotated datasets they
produce are, by any measure, very expensive. The result is that even
the largest annotated clinical text datasets are small by comparison
with those from other areas of NLP, and this has profound consequences
for the kinds of models that are viable in this space.

In this paper, we define a hybrid LSTM-CRF model that is effective
for real-world clinical text datasets. The architecture is sketched in
\figref{fig:model}. Its crucial property is that it synthesizes two
kinds of feature representation: dense representations that can be
trained on any large text corpus (not necessarily using clinical text)
and sparse, high-dimensional feature representations based on
hand-built feature functions. Hand-built feature functions are
especially powerful in healthcare because they can leverage the
numerous high-quality medical lexicons and ontologies that are
publicly available. As a result, such features can achieve impressive
coverage with relatively little additional effort.

We show that this combined model yields superior performance on five
datasets of diverse kinds of healthcare text: two clinical, one social
media, one scientific, and one commercial/regulatory (official drug labels).
Each task involves the labeling of complex, multi-word spans that pick
out diverse healthcare concepts: the Chemical--Disease Relation
dataset (CDR; \citealt{wei2015overview}); the Penn Adverse Drug
Reaction Twitter dataset (ADR; \citealt{Nikfarjam-etal:2015}); a new
disease diagnosis dataset; a new prescription reasons dataset that
involves identifying complex \nlplabel{Reason} spans for
drug--prescription actions; and a new dataset of 10K drug--disease
treatment descriptions, which we release with this paper.


\section{Models}\label{sec:models}

Our full model is depicted schematically in \figref{fig:model}. Its
modular structure defines a number of variations that allow us to
quantify the value of including dense and sparse feature
representations obtained from diverse sources.

Individual words are represented in two ways in the full model: with
dense, pretrained vectors and with sparse, high-dimensional feature
representations derived from hand-built feature functions. If the
dense representations are removed, the LSTM cells are also removed,
resulting in a standard CRF
\citep{Lafferty:McCallum:Pereira:2001,Stutton:McCallum:2010}. If the
sparse representations are removed, the result is a standard
LSTM-based RNN \citep{Hochreiter:Schmidhuber:1997}.

We explore two ways of initializing the dense representations: random
initialization according to the method of \citet{glorot2010} and the
ELMo embeddings released by \citet{peters2018deep}.  The ELMo
embeddings were trained on the 1 billion word benchmark of
\citet{chelba2013one} -- general newswire text not specialized to the
healthcare space. What is special about ELMo embeddings, as compared
to more standard word representation learning, is that they are
obtained from the parameters of a full language model, so that each
word's representation varies by, and is sensitive to, its linguistic
context; see also \citealt{McCann-etal:2017,radford2018improving}.

The nature of the hand-built feature representations varies by task,
so we leave most of the details to \secref{sec:experiments}. All the
models featurize each word in part using the word and part-of-speech
tag of the current word and the preceding and following four
words. They also include features that seek to characterize the nature
of the semantic environment: markers of negation, uncertainty,
hedging, and other core task-specific contextual cues. Finally, the
feature functions make extensive use of drug and disease lexicons to
identify the types of words. The drug lexicons are RxNorm, the
National Drug Code (NDC), FDA Drug Labels, FDA Orange Book, and the
OpenFDA fields found in a number of public FDA datasets (e.g., Drug
Adverse Events). The disease lexicons are derived from historical
ICD-9 and ICD-10 code sets, SNOMED-CT \citep{spackman1997snomed}, the
Disease Ontology \citep{schriml2011disease,kibbe2014disease}, and the
Wikidata graph \citep{vrandevcic2014wikidata}. The wealth and
diversity of these sources is typical of healthcare and highlights the
potential for taking advantage of such resources to help overcome the
challenges of small datasets. \Tabref{tab:featurization} shows an
 example of hand-built features. 

\newcommand{\dtag}[1]{\roamdarkblue{/#1}}
\newcommand{\other}{ }
\newcommand{\disease}{\dtag{\nlplabel{Disease} }}
\newcommand{\drug}{\dtag{\nlplabel{Drug} }}

\newcommand{\positive}{\dtag{\nlplabel{Positive} }}
\newcommand{\concern}{\dtag{\nlplabel{Concern} }}
\newcommand{\ruledout}{\dtag{\nlplabel{Ruled-out} }}

\newcommand{\reason}{\dtag{\nlplabel{Reason} }}
\newcommand{\prescribed}{\dtag{\nlplabel{Prescribed} }}

\newcommand{\adr}{\dtag{\nlplabel{ADR} }}

\newcommand{\treats}{\dtag{\nlplabel{Treats} }}
\newcommand{\contra}{\dtag{\nlplabel{Contra} }}

\begin{table*}[t]
  \centering
  \begin{tabular}[c]{T{0.21\linewidth} T{0.72\linewidth}}
    \toprule
    Dataset & Example \\
    \midrule
    Diagnosis  Detection
            & Asymptomatic\positive bacteriuria\positive ,\other
              could\other be\other neurogenic\concern bladder\concern
              disorder\concern .\other
    \\[2ex]
    Prescription Reasons
            & I will go ahead and place him on Clarinex\prescribed
              for\reason his\reason seasonal\reason allergic\reason
              rhinitis\reason .
    \\[2ex]
    Penn Adverse Drug Reactions (ADR)
            & \#TwoThingsThatDontMixWell venlafaxine and alcohol-
              you'll cry\adr and throw\adr chairs\adr at your mom's BBQ.
    \\[2ex]
    Chemical--Disease Relations (CDR)
            & Ocular\disease and\disease auditory\disease
              toxicity\disease in hemodialyzed patients receiving
              desferrioxamine\drug.
    \\[2ex]
    Drug--Disease Relations
            & Indicated for the management of active\treats
              rheumatoid\treats arthritis\treats and should not
              be used for rheumatoid\contra arthritis\contra in\contra
              pregnant\contra women\contra .
    \\
    \bottomrule
  \end{tabular}
  \caption{Short illustrative examples from each of our five datasets,
    with some modifications for reasons of space. CDR examples are
    typically much longer, encompassing an entire scientific title and
    abstract. \Secref{sec:experiments} more fully explicates the
    labels.  All unlabeled tokens are labeled with \nlplabel{Other}.}
  \label{tab:examples}
\end{table*}

In the full model, we include a dense layer that transforms the sparse
feature representations, and we apply dropout
\citep{Hinton-etal:2012} to this layer.  These transformed
representations are concatenated with the hidden states of the LSTM to
produce the full representations for each word. Where the hand-built
representations are left out, the word representations are simply the
hidden states of the RNN; where the dense representations are left
out, the word representations are simply the sparse representations,
resulting in a standard linear-chain CRF.

There is a natural variant of the model depicted in \figref{fig:model}
in which the CRF layer is replaced by a softmax layer. In our
experiments, this was always strictly worse than the CRF layer.
Another variant feeds the compressed hand-built features together with
ELMo embeddings into the LSTM. This too led to inferior or comparable 
performance. Finally, we evaluated a version that used a bidirectional
LSTM, but found that it did not yield improvements. Therefore, we
do not include those experimental results, to simplify the discussion.


\section{Experiments}\label{sec:experiments}

We report experiments on five different datasets: two from transcribed
clinical narratives, one from social media, one from scientific
publications, and one from official FDA Drug Labels texts. For each,
the task is to label spans of text that identify particular healthcare
concepts. We are particularly interested in the capacity of our models
to identify multi-word expressions in a way that is sensitive to the
semantics of the environment -- for example, to distinguish between a
drug prescribed and a drug discontinued, or to distinguish disease
mentions as diagnoses, diagnostic concerns, or ruled-out
diagnoses. \Tabref{tab:examples} gives a short illustrative example
from each dataset. \Tabref{tab:datastats} gives detailed
statistics for each dataset.

Three of the datasets are already partitioned into training and test
sets. For these, we tune the hyperparameters using 5-fold
cross-validation on the training set, train the model with tuned
hyperparameters on the training set, and then evaluate the performance
of the trained model on the test set.

The other two datasets do not have predefined splits. For these, we
divide them equally into five parts. For each fold, the
hyperparameters are tuned on the training data (also using 5-fold
cross-validation), and the best model is then applied to the test data
for the evaluation. These experiments are repeated three times to
smooth out variation deriving from the random initialization of the
model parameters, though we use the hyperparameters selected for each
fold in the first run in the subsequent two experiments to save
computational resources.

We use the Adam optimizer \citep{adam15}, with $\beta_{1} = 0.9$ and
$\beta_{2} = 0.999$, the training batch size set to $16$, and the dropout rate set to $0.5$ for all
the experiments. The step size $\eta$ and the coefficients of the
$\ell_{1}$ and $\ell_{2}$ regularizers $c_{1}$ and $c_{2}$ are tuned. The
step size is first tuned by setting both $c_{1} = c_{2} = 0$, and then
$c_{1}$ and $c_{2}$ are tuned using random search \citep{rnds} for ten
settings. \Tabref{tab:hyperparams} provides additional details on
our hyperparameters and evaluation protocol. The source code for our experiments and models is available.\footnote{\url{https://github.com/roamanalytics/roamresearch/tree/master/Papers/Feature4Healthcare}}

\begin{table*}[t]
  \centering
  \small
  \setlength{\tabcolsep}{6pt}
  \begin{tabular}[c]{*{6}{l}}
    \toprule
             & Diagnosis & Prescription &  Penn Adverse Drug  & Chemical--Disease &  Drug--Disease  \\
             & Detection & Reasons      &  Reactions (ADR)    & Relations (CDR)   &  Relations \\
    \midrule
    rand-LSTM-CRF    & 77.3 $\pm$ 0.05          & 69.6 $\pm$ 0.25          & 53.8 $\pm$ 0.88 & 85.1 $\pm$ 0.10 &  48.2 $\pm$ 1.12\\
    HB-CRF           & 82.0 $\pm$ 0.05          & 78.5 $\pm$ 0.01          &   58.8 $\pm$ 0.12    & 86.2 $\pm$ 0.02 & 42.3 $\pm$ 0.30\\
    ELMo-LSTM-CRF    & 83.9 $\pm$ 0.35          & 81.0 $\pm$ 0.20          &  65.7 $\pm$ 0.35 & 88.2 $\pm$ 0.34  & 50.6 $\pm$ 0.64 \\
    ELMo-LSTM-CRF-HB & \textbf{85.3} $\pm$ 0.24$^{***}$ & \textbf{82.0} $\pm$ 0.03$^{***}$ & \textbf{68.5} $\pm$ 1.67$^{*}$  & \textbf{89.9} $\pm$ 0.12$^{***}$ &  \textbf{51.9} $\pm$ 0.52$^{**}$ \\
   \bottomrule
  \end{tabular}
  \caption{Per-token macro-F1 scores. For ADR, the F1 scores are for
    chunks via approximate matching \citep{Nikfarjam-etal:2015,
      Tsai2006}.  `rand-LSTM' is an LSTM with randomly initialized
    word vectors. `ELMo-LSTM' is an LSTM initialized with pretrained
    ELMo embeddings. `HB' signals sparse, high-dimensional feature
    representations based on hand-built feature functions. The mean
    values and standard deviations are calculated using F1 scores of
    three runs of repeated experiments, as discussed in
    \secref{sec:experiments}. Statistical significance notation for the last two rows (two top-performing models) is $^*$: $p < 0.05$; $^{**}$: $p < 0.01$; $^{***}$: $p <0.001$.
}
  \label{tab:results}
\end{table*}


\subsection{Diagnosis Detection}\label{sec:diagnosis}

Our Diagnosis Detection dataset is drawn from a larger collection
clinical narratives -- de-identified transcriptions of the reports
healthcare professionals record about their interactions with
patients. The corpus was provided to us by a healthcare start-up.  We
sampled and labeled 6,042 sentences for information about disease
diagnoses. The labels are \nlplabel{Positive diagnosis},
\nlplabel{Concern}, \nlplabel{Ruled-out}, and \nlplabel{Other}.  The
labeling was done by a team of domain experts. The challenging aspects
of this task are capturing the complex, multi-word disease names and
distinguishing the semantic sense of those mentions (as summarized by
our label set) based on their sentential context.

For the hand-built parts of our representations, we extend the basic
feature set described in \secref{sec:models} with cue words that help
identify whether a description is about a patient's history or current
condition, as well as cue words for causal language, measurements, and
dates. The power these features bring to the model, beyond what is
captured in the ELMo-LSTM representations, is evident in
\tabref{tab:results}, column 1.


\subsection{Prescription Reasons}\label{sec:rx}

Our Prescription Reasons dataset is drawn from the same corpus of
clinical narratives as our Disease Diagnosis dataset and was annotated
by the same team of domain experts. This dataset contains 5,179
sentences, with labels \nlplabel{Prescribed}, \nlplabel{Discontinued},
\nlplabel{Reason}, and \nlplabel{Other}. For the first two labels, the
majority are unigrams naming drugs. Of special interest is the
\nlplabel{Reason} category, which captures long, highly diverse
reasons for actions taken concerning prescription drugs. (The relations are
captured with additional edge annotations connecting spans, but we do
not model them in this paper.)  This information about the rationale
for prescription decisions is the sort of thing that appears only in
text, and it has clear value when it comes to understanding these
decisions, making this an especially interesting task.

Our hand-built feature representations are similar to those used for
Diagnosis Detection, but they additionally contain features based in
large drug lexicons, as discussed in \secref{sec:models}, as well as
features based on cue-words for different prescription actions:
switching, discontinuing, increasing, decreasing, and so forth. The
results in \tabref{tab:results}, column 2, clearly favor the combined
model that uses both these features and the ELMo-LSTM.


\subsection{Penn Adverse Drug Reactions (ADR)}\label{sec:adr}

The Penn Adverse Drug Reactions (ADR; \citealt{Nikfarjam-etal:2015})
dataset is an annotated collection of tweets giving informal adverse
reactions to prescription drugs. It's thus a different kind of
clinical text than in our two previous experiments -- public
self-reports by patients, rather than private technical descriptions
by healthcare professionals.

The original dataset contained 1,340 labeled tweets for training and
444 for testing. However, due to restrictions on redistributing
Twitter data, the project team was unable to release the tweets, but
rather only a script for downloading them. Due to tweet deletions, we
were able to download only 749 train examples and 272 test
examples. This limits our ability to compare against prior work on
this dataset, but the small size further tests our hypothesis that our
combined model can get traction with relatively few examples.

For our hand-built feature functions, we follow the protocol specified
in the ADRMine CRF package released by \citet{Nikfarjam-etal:2015}.
Key components include tokenization
\citep{Gimpel:2011:PTT:2002736.2002747}, spelling correction
\citep{lucene, scowl}, lemmatization, and featurization
\citep{Loper:2002:NNL:1118108.1118117}. Thus our combined model is a
strict extension of this publicly available package (setting aside
differences related to implementation and optimization). We follow
\citet{Nikfarjam-etal:2015} in using Inside/Outside/Beginning (IOB;
\citealt{Ramshaw:Marcus:1995}) tags.

Our test-set results, given in \tabref{tab:results}, column 3, show
the power of our combined model. For context, the best results
reported by \citeauthor{Nikfarjam-etal:2015} are $72.1$, for a CRF
that includes hand-built features as well as features based on the
cluster indices of distributional word representations. That is, their
model draws on similar insights to our own. Though we only have half
of the training samples, our unified model is still able to get
traction on this dataset.


\subsection{Chemical--Disease Relations (CDR)}\label{sec:cdr}

The Biocreative V Chemical Disease Relation dataset of
\citet{wei2015overview} captures relationships between chemicals and
diseases in the titles and abstracts for scientific publications.  It
contains 1,000 training texts and 500 test texts. Its labels are
\nlplabel{Chemical}, \nlplabel{Disease}, and \nlplabel{Other}.  This
dataset is not only from a different domain than our others, but it
also involves much longer texts.

Our hand-built feature function is exactly the one used for the
Prescription Reasons experiments. We report results for the standard
test set. The power of the combined model is again evident in the
results in \tabref{tab:results}, column 4.


\subsection{Drug--Disease Relations}\label{sec:drug-disease}

Our final experiments are on a new annotated dataset that we will be
releasing along with this paper.\footnote{\url{https://github.com/roamanalytics/roamresearch/tree/master/BlogPosts/Features_for_healthcare}}
The underlying corpus is FDA Drug
Labels, which contains all the official labels for all drugs licensed
for sale in the U.S. These labels include a wide range of information,
including active ingredients, warnings, and approved usages.  Our
annotation project focused on capturing the relationship between these
drugs and mentioned diseases. The resulting labels are
\nlplabel{Treats}, \nlplabel{Prevents}, \nlplabel{Unrelated} and
\nlplabel{Contraindicated-for}. \Figref{fig:schemafda} describes the
corpus-building process in more detail.

Since FDA Drug Labels is a public dataset, we used this as an
opportunity to see whether we could obtain good labels via
crowdsourcing. This effort proceeded in two phases. In the first,
annotators identified disease spans, working from an annotation manual
that provided guidance on how to delimit such phrases and lexical
resources to help them identify diseases. In the second phase,
annotators assigned the span labels from our label set, again using an
annotation manual we created to guide their choices.

We launched our task on Figure Eight with 10,000 sentences. It was
completed within a few days. The job was done by 1,771 people from 72
countries, the majority from Venezuela. No special qualifications were
imposed. To infer a label for each example, we applied Expectation
Maximization (EM), essentially as in \citet{dawid1979maximum}. The
inter-annotator agreement between these labels and those we inferred
via EM is $0.83$ for both tasks. For assessment, a team of experts
independently labeled 500 examples from the same pool of sentences,
using the same criteria and annotation manuals as the
crowdworkers. The inter-annotator agreement between the labels
inferred from the crowd and those from the experts is $0.82$,
suggesting that the inferred labels are good.

We expect the crowdsourced labels to be used only for training.  Our
test set consists entirely of non-train examples with labels assigned
by experts. This allows us to train on noisy labels, to check for
robustness, while still assessing on truly gold labels. Our results
for this experiment are given in \tabref{tab:results}, column 5, and
point to the superiority of our combined model.


\section{Discussion}\label{sec:discussion}

Our discussion seeks to show that the combined model, which shows
superior performance in all tasks (\tabref{tab:results}), is making
meaningful use of both kinds of features (hand-built and ELMo) and both
of the major model components (LSTM and CRF).

\subsection{The Role of Text Length}

\begin{figure*}[t]
  \centering
  \includegraphics[width=1.0 \linewidth]{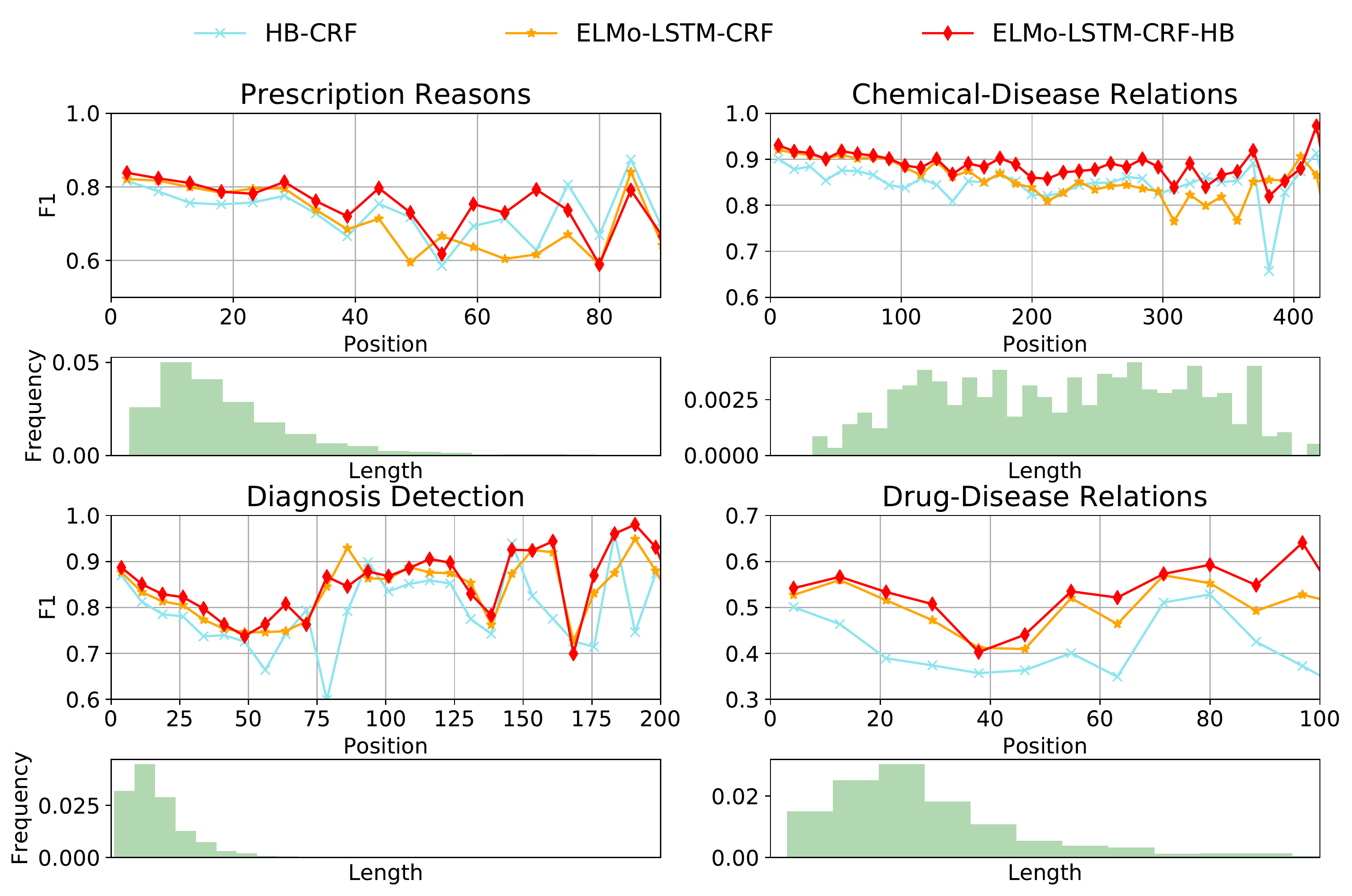}
  \caption{Text-length experiments.  Along with the distribution of
    text lengths, per-token macro-F1 scores of words that fall into
    specific bins in the sentences are shown.  For the top two
    datasets, the ELMo-LSTM-CRF is better at earlier positions, while
    the HB-CRF is better at later ones.  For the bottom two datasets,
    the ELMo-LSTM-CRF is always better than the HB-CRF. In all these
    cases, the combined model takes advantage of both models and
    always outperforms the base models. 
    ADR dataset results are given in 
    \figref{fig:lengthsadr}
    due to space limitations.}
  \label{fig:lengths}
\end{figure*}

We expect the LSTM to handle short texts very effectively, but that
its performance will be degraded for long ones. In contrast, the CRF
might fall short of the LSTM on short texts, but it should be more
robust on long ones. We thus hypothesize that the combined model will
learn to take advantage of these comparative strengths.

We find strong support for this hypothesis in our data.
\Figref{fig:lengths} illustrates this. These plots track the macro-F1
scores (y-axes) of tokens in specific linear positions (x-axes).
There are two major trends.

First, in the Prescription Reasons and CDR datasets (top two panels),
we see that the HB-CRF starts to outperform the ELMo-LSTM-CRF after
about word 40 in Prescription Reasons (which contains many long texts that list patient history; \secref{sec:adr}) and after about word 160 in CDR
(which has paragraph-length texts; \secref{sec:cdr}).

Second, in the Diagnosis Detection and Drug--Disease Relations
datasets (bottom two panels in \figref{fig:lengths}), the
ELMo-LSTM-CRF model outperforms the HB-CRF at all positions. However,
there is still evidence that our full model is leveraging the
strengths of both of its major components, as it outperforms both in
all positions.

In summary, the performance curve of the combined model is roughly an
upper envelope of the two base-model curves. The combined model is
able to achieve better performance for both short and long texts, and
for words in any position, by utilizing features from both base
models.

\subsection{Analysis of the CRF Potential Scores}

The potential scores (also referred to as  ``unary scores'' or ``emissions'' in some work) of the CRF provide another method for model
introspection. These scores are the direct inputs to the final CRF
layer, where the token-level label predictions are determined. When
the potential score for a specific label is high, the CRF assigns a
high weight to that label under the contraints of adjacent
labels. Thus, by checking the potential scores for the feature
dimensions deriving from each of our base models, we can gain insights
into the relative importance of these models and how the combined
model leverages features from both.

\begin{figure*}[t]
  \centering
  \includegraphics[width=1.0\linewidth]{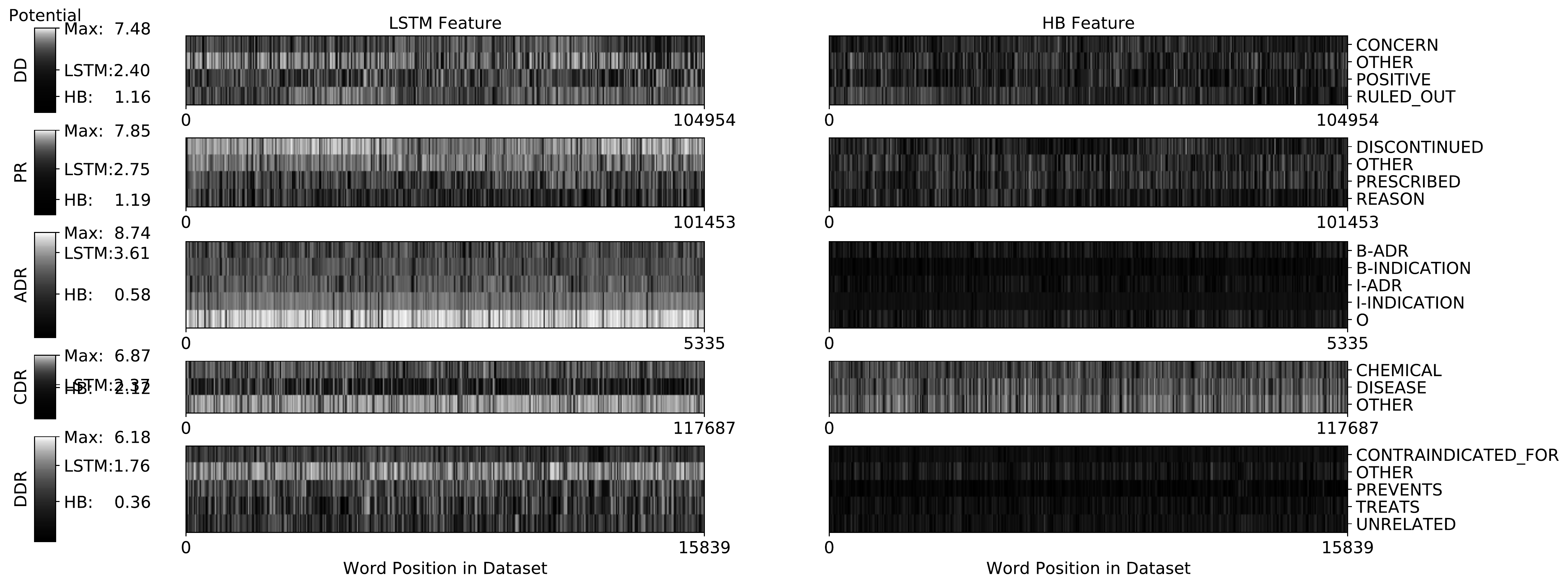}
  \caption{Potential score experiments. Potential scores from the 
    ELMo-LSTM and HB modules of all five datasets are shown. Mean
    potential scores of both features are calibrated in the left
    colorbar. Higher potential scores (lighter cells) indicate greater
    importance for the feature. In all five datasets, the combined
    model pays more attention to the ELMo-LSTM features, but the
    hand-built features always contribute. Comparing with the results
    in \tabref{tab:results}, we note that when the performance of two
    base models is comparable, their potential scores in the combined
    model are also closer.}
  \label{fig:potentials}
\end{figure*}

The potential scores of each word in the test set are shown in 
\figref{fig:potentials}, where the left panels show the LSTM features
and the right panels show the CRF (hand-built) features. Due to the
general effectiveness of the ELMo-LSTM, we always have higher average
potential scores from those features. This is reflected in the mean
scores at left and in the comparatively large amount of white (high
scores) in the panels. However, the hand-built features always make
substantial contributions, especially in Diagnosis Detection,
Prescription Reasons, and CDR. We note also that, where the
performance of the two base models is very similar
(\tabref{tab:results}), the potential scores in the combined model are
also more similar.

\subsection{Major Improvements in Minor Categories}

One of our central motivations for this work is that clinical datasets
tend to be small due to the challenges of getting quality labels on
quality data. These size limitations impact model performance, and the
hardest hit categories tend to be the smallest ones. Unfortunately,
these are often the most important categories, identifying rare but
significant events. We are thus especially interested in whether our
combined model can address this problem.

\Tabref{tab:improvement} suggests that the combined model does make
progress here, in that the largest gains, across all relevant
datasets, tend to be for the smallest categories. This is very
dramatically true for the Drug--Disease Relations dataset, where only
the combined model is able to get any traction on the smallest
categories; it achieves $103.5\%$ and $71.3\%$ improvements in F1
score over the HB-CRF model for the two smallest categories. It seems
clear that, in transferring compact embedding representations
learned from other large text datasets, the combined model can elevate
performance on small categories to an acceptable level.

\begin{table*}[t]
  \centering
  \small
  \setlength{\tabcolsep}{6pt}
  \begin{tabular}[c]{*{8}{r}}
    \toprule
    \multicolumn{4}{r}{Diagnosis Detection}  & \multicolumn{4}{r}{Prescription Reasons} \\
    Label & Support & F1 score  & Improvement       &               Label & Support  &F1 score & Improvement \\
    \midrule
    \nlplabel{Other} & 74888 & 95.3 & 1.4\%        &             \nlplabel{Other}  & 83618  & 95.8 & 0.9\% \\
    \nlplabel{Positive}  & 24489 & 86.1 & 4.4\%        &              \nlplabel{Reason} & 9114  & 64.7 & 8.6\% \\
    \nlplabel{Ruled-out} & 2797 & 86.4 & 3.6\%        &              \nlplabel{Prescribed} & 5967  & 84.7 & 4.4\% \\
    \nlplabel{Concern} & 2780 & 72.1 & 5.6\%        &              \nlplabel{Discontinued} & 2754  & 82.7 & 5.6\% \\
    \midrule
    \multicolumn{4}{r}{Chemical--Disease Relations (CDR)}  & \multicolumn{4}{r}{Drug--Disease  Relations} \\
    Label & Support  & F1 score & Improvement       &               Label & Support& F1 score  & Improvement \\
    \midrule
    \nlplabel{Other}  & 104530 & 98.3 & 0.5\%        &             \nlplabel{Other}  & 10634 & 90.8 & 2.3\% \\
    \nlplabel{Disease} & 6887 & 84.2  &6.3\%        &              \nlplabel{Treats} & 3671& 76.0  & 5.7\% \\
    \nlplabel{Chemical} & 6270& 87.0 & 6.7\%        &              \nlplabel{Unrelated} & 1145 & 53.8 & 71.3\% \\
     &  &     &    &              \nlplabel{Prevents} & 320 & 41.1 & 103.5\% \\
     &  &     &    &              \nlplabel{Contraindicated-for} & 69& 0  & -- \\
    \bottomrule
  \end{tabular}
  \caption{Relative F1 score improvements of different labels.  For
    each label, we give the number of supporting examples (Support),
    the F1 score of our combined model, and the relative improvements
    over the HB-CRF model. The F1 scores of minor labels suffer from
    insufficient training data, and thus have lower values. However,
    the combined model shows the largest relative improvements in
    these categories. 
    ADR results are shown in 
    \tabref{tab:improvementadr}.}
  \label{tab:improvement}
\end{table*}


\section{Prior Work}\label{sec:lit}

\subsection{Clinical Text Labeling}\label{sec:clinical-text}

Apache cTAKEs \citep{doi:10.1136/jamia.2009.001560} 
extracts information from clinical text. Its labeling module
implements a dictionary look-up of concepts in the UMLS database, and
the concept is then mapped into different semantic types
(labels). Similar extractions play a role in our hand-built features,
but only as signals that our models learn to weight
against each other to make decisions.

ADRMine \citep{Nikfarjam-etal:2015} is closer to our own approach; it
focuses on extracting adverse drug reaction mentions from noisy
tweets. It combines hand-built features and word embedding cluster
features for label prediction. However, our model is more powerful in
the sense that we directly utilize the word embeddings and feed them
into the LSTM.

\citet{doi:10.1093/bioinformatics/btx228} use a combined LSTM-CRF to
achieve better NER results on 33 biomedical datasets than both
available NER tools and entity-agnostic CRF methods, though they
do not incorporate hand-built features.

There are also competitions related to labeling tasks in the context
of clinical text. The i2b2 Challenge
\citep{doi:10.1136/amiajnl-2013-001628} includes event detection as
one of the task tracks, which is basically a labeling task. The best
results on this task came from a team using a simple CRF. The
Biocreative V Chemical--Disease relation (CDR) competition
\citep{wei2015overview} released a widely used dataset for researchers
to evaluate their NER tools for biomedical text, and \citet{Verga}
report state-of-the-art results for a self-attention encoder, using a
dataset that extends CDR.

\subsection{Efficient Annotation}\label{sec:efficient-annotation}

Obtaining accurate annotations is expensive and time consuming in many
domains, and a rich line of research seeks to ease this annotation
burden. \citet{NIPS2016_6523} and \citet{P18-1175} propose to
synthesize noisy labeling functions to infer gold training labels, and
thus make better use of annotators' time, by allowing them to focus on
writing high-level feature functions (and perhaps label individual
examples only for evaluation). These efforts are potentially
complementary to our own, and our experiments on our new Drug--Disease
dataset (\secref{sec:drug-disease}) suggest that our combined model is
especially robust to learning from noisy labels compared with base models.

\subsection{Related Models}\label{sec:related-models}

A large body of work explores combined LSTM and CRF models for text
labeling. \citet{Huang2015BidirectionalLM} use an LSTM-CRF for
sequence tagging, and \citet{Ma2016EndtoendSL} propose a 
bi-directional LSTM-CNNs-CRF for the same task. In addition to word
embeddings, \citet{Lample2016NeuralAF} utilize character embedding
information as the input to a LSTM-CRF.
\citet{Jagannatha2016StructuredPM} integrate pairwise potentials into
the LSTM-CRF model, which improves sequence-labeling performance in
clinical text.
\citet{doi:10.1093/bioinformatics/bty869} and \citet{Crichton2017} use
multi-task learning based on the basic LSTM-CRF structure to improve NER
performance in biomedical text. Our model provides an effective method
for fully utilizing the sparse ontology-driven features left out of
by the above work, which are complementary to dense embeddings and therefore boost performance of clinical concept extraction with limited training data (\secref{sec:discussion}).

There are also a number of models that mix dense and sparse feature
representations. \citet{Gormley_2015b} and \citet{cheng2016widedeep} combine both unlexicalized hand-crafted features and word embeddings to improve the performance of relation extraction in recommender systems. However, they focus on simple multi-layer perceptron models, rather than considering a more expressive LSTM structure. Similarly, \citet{Wang434803} utilize both sparse UMLS features and unpretrained word embeddings as the input to an LSTM for genetic association inferences from medical literature. While their UMLS features are a single look-up table of semantic types, our model relies on much richer resources of medical knowledge and includes more heterogeneous and expressive hand-built features that capture the semantic, morphological and contextual information of words (\secref{sec:models}).


\section{Conclusion}\label{sec:conclusion}

Clinical text datasets are expensive to label and thus tend to be
small, but the questions they can answer are often very
high-impact. It is thus incumbent upon us to make maximally efficient
use of these resources. One way to do this is to draw heavily on
lexicons and other structured resources to write feature
functions. Another way is to leverage unlabeled data to create dense
feature vectors.

The guiding hypothesis of this paper is that the best models will make
use of both kinds of information. To explore this hypothesis, we
defined a new LSTM-CRF architecture that brings together these two
kinds of feature, and we showed that this combined model yields
superior performance on five very different healthcare-related tasks.
We also used a variety of introspection techniques to gain an
understanding of how the combined model balances its different sources
of information. These analyses show that the combined model learns to
pay attention to the most reliable sources of information for
particular contexts, and that it is most effective, as compared to its
simpler variants, on smaller categories, which are often the most
crucial and the hardest to generalize about.

We also introduced the publicly available Drug--Disease Relations
dataset, which contains a large training set of crowdsourced labels
and a smaller test set of gold labels assigned by experts. This
dataset can be used to learn facts about drug--disease relationships
that have medical significance, and it shows that combined models like
ours can learn effectively in noisy settings.


\bibliography{features-for-healthcare-bib}
\bibliographystyle{acl_natbib}

\setcounter{table}{0}
\renewcommand{\thetable}{A\arabic{table}}

\setcounter{figure}{0}
\renewcommand{\thefigure}{A\arabic{figure}}


\appendix



\begin{table*}[t]
  \setlength{\tabcolsep}{4pt}
	\centering
	\begin{tabular}[c]{@{} T{0.1\linewidth} T{0.88\linewidth} @{}}
		\toprule
		Sentence & Hand-built features of word \word{\textbf{bacteria}} \\
		\midrule
antiseptic handwash to decrease \word{\textbf{bacteria}}
on \ \ \ \ \ \ \ \
the \ \ \ \ \ 
skin \ \ \ \ \ \ \ \ \ 
. & 
\textbf{Adjacent words features}:  \ \ \ \ \ \ \ \ \ \ \ \ \ \ \ \ \ \ \ \ \ \ \ \ \ \ \ \ \ \ \ \ \ \ \ \ \ \ \ \ \ \ \ \ \ \ \ \ \ \ \ \ \ \ \ \ \ \ \ \ \ \ \ \ \ \ \ \ \ \ \ \ \ \   \ \ \ \ \ \ \ \ \ \ \ \ \ \ \ \ \ \ \ \ \ \ \ \ \ \ \ \ \ \ \ \ \ \ \ \ \ \ \ \ \ \ \ \ \ \ \ \ \ \ \ \ \ \ \ \ \ \ \ \ \ \ \ \ \ \ \ \ \ \ \ \ \ \ 
word-4:antiseptic, word-3:handwash, word-2:to, word-1:decrease, \ \ \ \ \ \ \ \ \ \ \ \ \ \  \ \ \ \ \ \ \ \ \ \ \ word:bacteria, 
word+1:on, word+2:the, word+3:skin, word+4:.. \ \ \ \ \ \ \ \ \ \ \ \ \ \ \ \ \ \ \ \ \ \ \ \ \ \ \ \ \ \ \ \ \ \ \ \ \ 
\textbf{Adjacent POS tags features}:  \ \ \ \ \ \ \ \ \ \ \ \ \ \ \ \ \ \ \ \ \ \ \ \ \ \ \ \ \ \ \ \ \ \ \ \ \ \ \ \ \ \ \ \ \ \ \ \ \ \ \ \ \ \ \ \ \ \ \ \ \ \ \ \ \ \ \ \ \ \ \ \ \ \   \ \ \ \ \ \ \ \ \ \ \ \ \ \ \ \ \ \ \ \ \ \ \ \ \ \ \ \ \ \ \ \ \ \ \ \ \ \ \ \ \ \ \ \ \ \ \ \ \ \ \ \ \ \ \ \ \ \ \ \ \ \ \ \ \ \ \ \ \ \ \ \ \ \ 
tag-4:JJ, tag-3:NN, tag-2:TO, tag-1:VB,  \ \ \ \ \ \ \ \ \ \ \ \ \ \ \ \ \ \ \ \ \ \ \ \ \ \ \ \ \ \ \ \ \ \ \ \ \ \ \ \ \ \ \ \ \ \ \ \ \ \ \ \ \ \ \ \ \ \ \ \ \ \ \ \ \ \ \ \ \ \ \ \ \ \ 
tag:NNS, tag+1:IN, tag+2:DT, tag+3:NN, tag+4:.. \ \ \ \ \ \ \ \ \ \ \ \ \ \ \ \ \ \ \ \ \ \ \ \ \ \ \ \ \ \ \ \ \ \ \ \ \ \ \ \ \ \ \ \ \ \ \ \ \ \ \ \ \ \ \ \ \ \ \ \ \ \ \ \ \ \ \ \ \ \ \ \ \ \
\textbf{Semantic environment features}:  \ \ \ \ \ \ \ \ \ \ \ \ \ \ \ \ \ \ \ \ \ \ \ \ \ \ \ \ \ \ \ \ \ \ \ \ \ \ \ \ \ \ \ \ \ \ \ \ \ \ \ \ \ \ \ \ \ \ \ \ \ \ \ \ \ \ \ \ \ \ \ \ \ \   \ \ \ \ \ \ \ \ \ \ \ \ \ \ \ \ \ \ \ \ \ \ \ \ \ \ \ \ \ \ \ \ \ \ \ \ \ \ \ \ \ \ \ \ \ \ \ \ \ \ \ \ \ \ \ \ \ \ \ \ \ \ \ \ \ \ \ \ \ \ \ \ \ \  
bias:1, is\_upper:0, is\_title:0, is\_punctuation:0, \ \ \ \ \ \ \ \ \ \ \ \ \ \ \ \ \ \ \ \ \ \ \ \ \ \ \ \ \ \ \ \ \ \ \ 
in\_left\_context\_of\_negative\_cues:0, \ \ \ \ \ \ \ \ \ \ \ \ in\_right\_context\_of\_negative\_cues:0, 
in\_left\_context\_of\_prevents\_cues:0, \ \ \ \ \ \ \ \ \ \ \ \ in\_right\_context\_of\_prevents\_cues:0, 
in\_left\_context\_of\_treats\_cues:0, \ \ \ \ \ \ \ \ \ \ \ \ \ \ \ \ \ in\_right\_context\_of\_treats\_cues:0,
in\_left\_context\_of\_treats\_symptoms\_cues:0, in\_right\_context\_of\_treats\_symptoms\_cues:0,  in\_left\_context\_of\_contraindicated\_cues:0, \  in\_right\_context\_of\_contraindicated\_cues:0, in\_left\_context\_of\_affliction\_adj\_cues:0, \ \ \ \ \  in\_right\_context\_of\_affliction\_adj\_cues:0, in\_left\_context\_of\_indication\_cues:0, \ \ \ \ \ \ \ \ \ \ in\_right\_context\_of\_indication\_cues:0, 
in\_left\_context\_of\_details\_cues:0, \ \ \ \ \ \ \ \ \ \ \ \ \ \ \ \
in\_right\_context\_of\_details\_cues:0. \ \ \ \ \ \ \ \ \ \ \ \  \ \ \ \ \ \ \ \ \ \ \ \
 \\
		\bottomrule
	\end{tabular}
	\caption{Hand-built features of the word \word{\textbf{bacteria}} in a Drug--Disease Relations dataset example. These features describe the word's adjacent words, adjacent POS tags, and semantic environment (\secref{sec:models}).
	The detailed meanings of hand-built features in the table are described as below: \textbf{Adjacent words features}: ``word($\pm$1/2/3/4)'' feature the word and adjacent words within a window size of 9. \textbf{Adjacent POS tags features}: ``tag($\pm$1/2/3/4)'' feature the tags of word and its adjacent words within a window size of 9.
	\textbf{Semantic environment features}:
	``bias'' is always 1 for all words; ``is\_upper'' specifies whether the word is upper case or lower case; ``is\_title'' features whether the word is in the title or not; ``is\_punctuation'' specifies whether the token is actually a word or a punctuation. ``in\_left/right\_context\_of\_negative/prevents/treats(\_symptoms)/contraindicted/afflicition\_adj/indication/details\_cues'' feature whether the word is in the left or right context (of specific window size like 4) of cue-words from specific lexicons. Features related to 8 lexicons are shown in this example. Concrete examples: \word{not}, \word{none} and \word{no} are three cue-words of lexicon ``negative\_cues'', \word{prevent} and \word{avoid} are two cue-words of lexicon ``prevents\_cues'', \word{treat}, \word{solve} and \word{alleviate} are three cue-words of lexicon ``treats\_cues'' etc.
	Different semantic environments are defined in the five datasets by carefully defining the lexicons/cue-words from various sources which possibly contain corresponding domain knowledge, as discussed in \secref{sec:models} and \secref{sec:experiments}. 
	}
	\label{tab:featurization}
\end{table*}

\begin{table*}[t]
	\centering
	\small
	\setlength{\tabcolsep}{6pt}
	\begin{tabular}[c]{*{6}{r}}
		\toprule
		& Diagnosis & Prescription &  Penn Adverse Drug  & Chemical--Disease &  Drug--Disease  \\
		Statistics & Detection & Reasons      &  Reactions (ADR)    & Relations (CDR)   &  Relations \\
		\midrule
		\# texts    & 6042          & 5179          &  --  & -- & -- \\
		\# training texts    & --          & --      & 749 & 1000 & 9494 \\
		\# test texts           & --         &--       &  272 & 500 & 500 \\
		mean text length    & 17          & 19 &  19 & 227 & 30 \\
		max text length & 374 &258 & 40 & 623 & 542 \\
		\# labels & 4 &4                           & 5 & 3 & 5 \\
		\bottomrule
	\end{tabular}
	\caption{Statistics for our five datasets. The sample size varies
		from around 1,000 to 10,000. The mean text length (measured as the number of words) varies from 17
		(short sentences) to 227 (full paragraphs). The number of labels
		varies from 3 to 5. ADR, CDR, and Drug--Disease Relations are
		already partitioned into training and test sets, while Diagnosis
		Detection and Prescription Reasons do not have predefined splits.}
	\label{tab:datastats}
\end{table*}

\begin{table*}[t]
  \centering
  \small
  \setlength{\tabcolsep}{4pt}
  \begin{tabular}[c]{@{} *{7}{r} @{}}
    \toprule
    &  & Diagnosis &  Prescription & Penn Adverse Drug & Chemical--Disease & Drug--Disease\\
    Models & Hyperparams & Detection      &  Reasons    & Reactions (ADR)   &  Relations (CDR) &Relations\\
    \midrule
    \multirow{5}{*}{rand-LSTM-CRF}    
    & $\eta$          &       1e-4         &   1e-4  &  1e-4 &  1e-4 &  1e-4 \\
    & $\textrm{epoch}_{\textrm{tune}}$         & 3          &  3  & 513 & 10 & 13 \\
    & $\textrm{epoch}_{\textrm{train}}$          & 34          &  40  & 3076 & 164 & 130 \\
    & $\mathcal{R}_{c1}$          &  \multicolumn{5}{c}{$\{$ 0, 3e-5, 1e-4, 3e-4, 1e-3 $\}$} \\
    & $\mathcal{R}_{c2}$          & \multicolumn{5}{c}{$\{$ 0, 3e-4, 1e-3, 3e-3, 1e-2 $\}$} \\
    \midrule
    \multirow{5}{*}{HB-CRF}
    & $\eta$    &1e-2      &   1e-2       &  3e-2  & 1e-2 & 1e-4 \\
    & $\textrm{epoch}_{\textrm{tune}}$         & 1           &  1  & 10 & 2 & 3 \\
    & $\textrm{epoch}_{\textrm{train}}$        & 3          &  4 & 82 & 10 & 35 \\
    & $\mathcal{R}_{c1}$          & \multicolumn{5}{c}{$\{$ 0, 3e-6, 1e-5, 3e-5, 1e-4 $\}$} \\
    & $\mathcal{R}_{c2}$          & \multicolumn{5}{c}{$\{$ 0, 3e-5, 1e-4, 3e-4, 1e-3 $\}$} \\
    \midrule
    \multirow{5}{*}{ELMo-LSTM-CRF}
    & $\eta$          & 1e-3    &  1e-3  & 1e-4 & 1e-3 & 5e-6 \\
    &$\textrm{epoch}_{\textrm{tune}}$      & 1           &  1  & 10 & 2 & 3 \\
    & $\textrm{epoch}_{\textrm{train}}$        & 3          &  4  & 82 & 10 & 35 \\
    & $\mathcal{R}_{c1}$          & \multicolumn{5}{c}{$\{$ 0, 3e-5, 1e-4, 3e-4, 1e-3 $\}$} \\
    & $\mathcal{R}_{c2}$          & \multicolumn{5}{c}{$\{$ 0, 3e-4, 1e-3, 3e-3, 1e-2 $\}$} \\
    \midrule
    \multirow{5}{*}{ELMo-LSTM-CRF-HB}
    & $\eta$          & 1e-3          & 1e-3  & 1e-4 & 1e-3 & 1e-5 \\
    & $\textrm{epoch}_{\textrm{tune}}$        & 1        &  1  & 10 & 2 & 3 \\
    & $\textrm{epoch}_{\textrm{train}}$        & 3          & 4  & 82 & 5 & 35 \\
    & $\mathcal{R}_{c1}$          & \multicolumn{5}{c}{$\{$ 0, 3e-7, 1e-6, 3e-6, 1e-5 $\}$} \\
    & $\mathcal{R}_{c2}$          & \multicolumn{5}{c}{$\{$ 0, 3e-6, 1e-5, 3e-5, 1e-4 $\}$} \\
    \bottomrule
  \end{tabular}
  \caption{Hyperparameters for our experiments. The step size $\eta$
    is first manually tuned within the training set when the $\ell_{1}$
    and $\ell_{2}$ regularizers are set to be zeros. The coefficients
    $c_1$ and $c_2$ of the $\ell_{1}$ and $\ell_{2}$ regularizers are
    determined via random search (for 10 settings) from ranges
    $\mathcal{R}_{c1}$ and $\mathcal{R}_{c2}$ during tuning \citep{rnds}.  Epochs of
    tuning $\textrm{epoch}_{\textrm{tune}}$ are set to 1$\sim$3 to
    reduce tuning time for most datasets (which consumes most of the
    time for the experiments). It is set to 10 for ADR since that
    dataset is so small that it is hard to see clear trends after just
    one epoch. Epochs of training $\textrm{epoch}_{\textrm{train}}$
    are set to be large enough until the training converges. The
    `rand-LSTM-CRF' model requires many more epochs for tuning and
    training because of the updates to the randomly initialized
    embeddings.}
  \label{tab:hyperparams}
\end{table*}

\begin{figure*}[t]
	\centering
	\includegraphics[width=1 \linewidth]{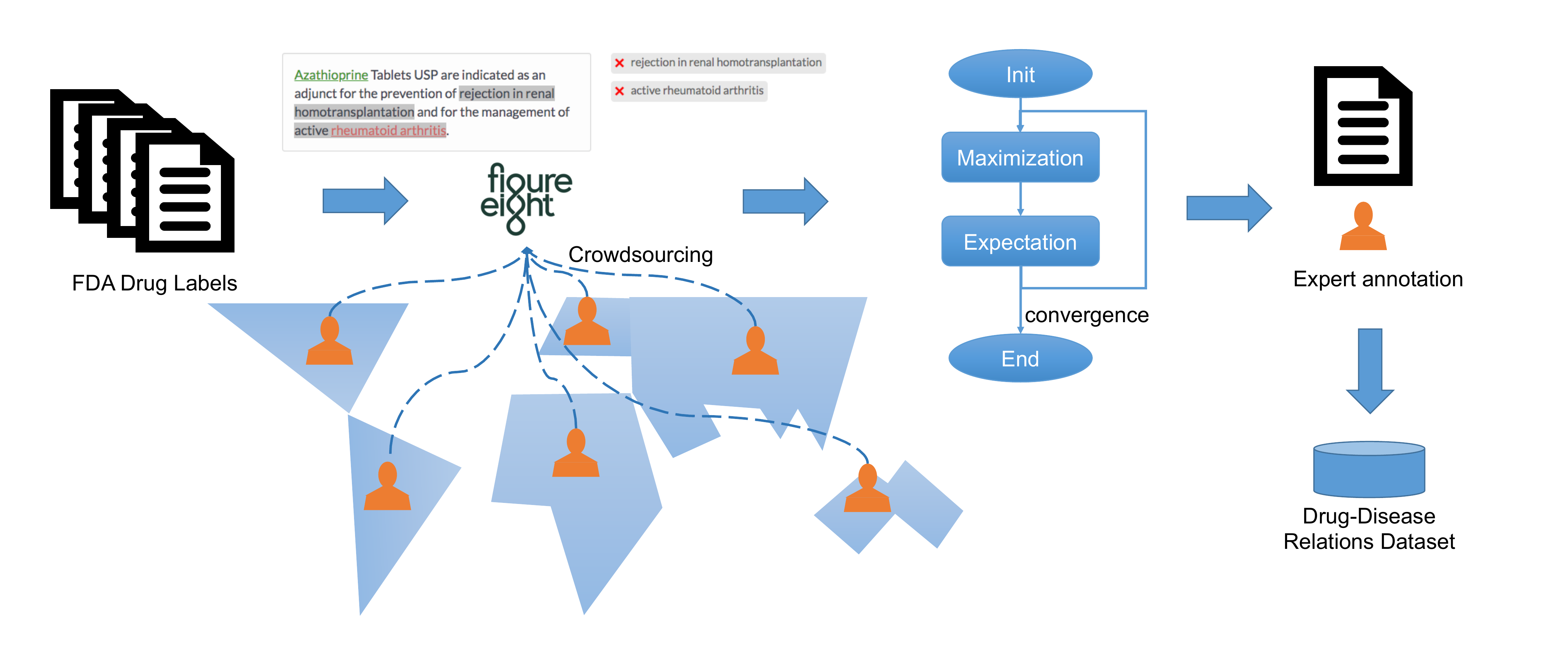}
	\caption{Procedure for building the Drug--Disease Relations dataset. 10,000 raw sentences from the FDA Drug Labels corpus were annotated by participants from 72 countries on the Figure Eight platform (crowdsourcing). Expectation Maximization was used to infer labels for all the annotated sentences used for training. A team of experts independently labeled different examples for testing. The resulting dataset consists of 9,500 crowdsourced examples and 500 expert-annotated examples.}
	\label{fig:schemafda}
\end{figure*}

\begin{figure*}[t]
	\centering
	\includegraphics[width=0.5 \linewidth]{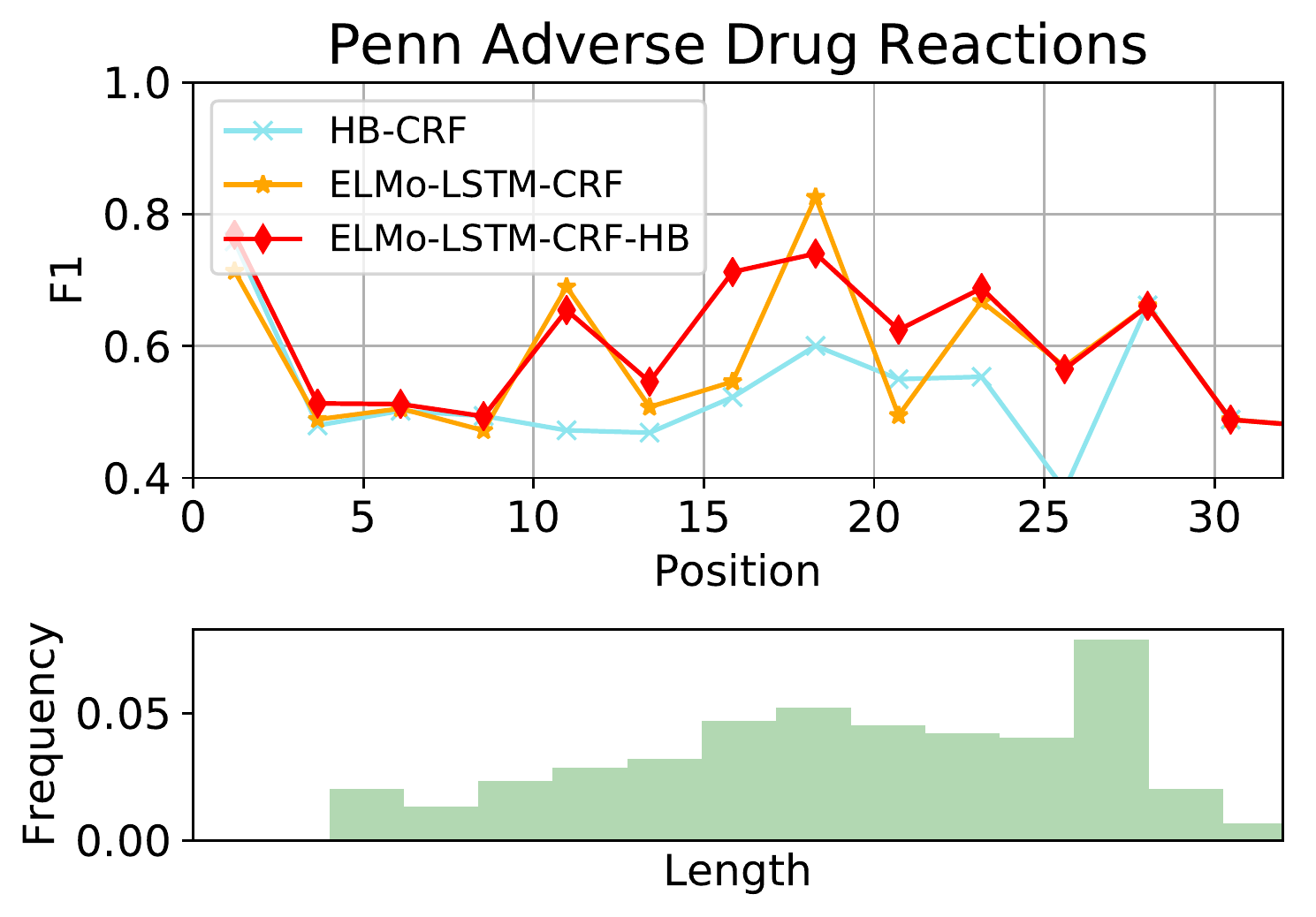}
	\caption{Text-length experiment for the Penn Adverse Drug Reactions (ADR) dataset. Since ADR uses the IOB tag format, in order to calculate per-token F1 scores, we collapse test-set labels starting with `B-' and `I-' into the same labels. The ELMo-LSTM-CRF always performs better than the HB-CRF, while the combined model takes advantage of both models and always outperforms both base models. \Figref{fig:lengths} provides comparable plots for the other four datasets.}
	\label{fig:lengthsadr}
\end{figure*}

\begin{table*}[t]
	\centering
	\setlength{\tabcolsep}{6pt}
	\begin{tabular}[c]{*{4}{r}}
		\toprule
		\multicolumn{4}{r}{Penn Adverse Drug Reactions (ADR)}   \\
		Label & Support & F1 score  & Improvement        \\
		\midrule
		\nlplabel{Other} & 5023 & 98.0 & 0.3\%    \\
		\nlplabel{ADR}  & 283 & 57.1 & 17.7\%    \\
		\nlplabel{Indication} & 29 & 35.9 & 178.3\%     \\
		\bottomrule
	\end{tabular}
	\caption{Relative F1 score improvements of different labels in the Penn Adverse Drug Reactions (ADR) dataset. To bring the IOB tag format of this dataset in line with our others, \nlplabel{ADR} merges \nlplabel{B-ADR} and \nlplabel{I-ADR}, and \nlplabel{Indication} merges \nlplabel{B-Indication} and \nlplabel{I-Indication}. Consistent with \tabref{tab:improvement}, the combined model gains most in the smallest categories.}
	\label{tab:improvementadr}
\end{table*}

\end{document}